\documentclass[10pt,twocolumn,letterpaper]{article}

\usepackage{icb}
\usepackage{times}
\usepackage{epsfig}
\usepackage{graphicx}
\usepackage{amsmath}
\usepackage{amssymb}
\usepackage{url}
\usepackage{subcaption}
\usepackage{booktabs}

\usepackage[boxruled]{algorithm2e}
\usepackage{siunitx}

\icbfinalcopy 

\ificbfinal\fi

\begin{document}

\title{Learning-Free Iris Segmentation Revisited: A First Step Toward Fast Volumetric Operation Over Video Samples}

\author{Jeffery Kinnison\\
University of Notre Dame\\
Notre Dame, Indiana, USA\\
{\tt\small jkinniso@nd.edu}
\and
Mateusz Trokielewicz\\
Research and Academic Computer Network\\
Warsaw, Poland\\
{\tt\small mateusz.trokielewicz@nask.pl}\\
\and
Camila Carballo\\
University of Notre Dame\\
Notre Dame, Indiana, USA\\
{\tt\small ccarball@nd.edu}\\
\and
Adam Czajka\\
University of Notre Dame\\
Notre Dame, Indiana, USA\\
{\tt\small aczajka@nd.edu}\\
\and
Walter Scheirer\\
University of Notre Dame\\
Notre Dame, Indiana, USA\\
{\tt\small walter.scheirer@nd.edu}\\
}

\maketitle
\thispagestyle{empty}

\begin{abstract}

Subject matching performance in iris biometrics is contingent upon fast, high-quality iris segmentation. In many cases, iris biometrics acquisition equipment takes a number of images in sequence and combines the segmentation and matching results for each image to strengthen the result. To date, segmentation has occurred in 2D, operating on each image individually. But such methodologies, while powerful, do not take advantage of potential gains in performance afforded by treating sequential images as volumetric data. As a first step in this direction, we apply the Flexible Learning-Free Reconstructoin of Neural Volumes (FLoRIN) framework, an open source segmentation and reconstruction framework originally designed for neural microscopy volumes, to volumetric segmentation of iris videos. Further, we introduce a novel dataset of near-infrared iris videos, in which each subject's pupil rapidly changes size due to visible-light stimuli, as a test bed for FLoRIN. We compare the matching performance for iris masks generated by FLoRIN, deep-learning-based (SegNet), and Daugman's (OSIRIS) iris segmentation approaches. We show that by incorporating volumetric information, FLoRIN achieves a factor of 3.6 to an order of magnitude increase in throughput with only a minor drop in subject matching performance. We also demonstrate that FLoRIN-based iris segmentation maintains this speedup on low-resource hardware, making it suitable for embedded biometrics systems.

\end{abstract}

\section{Introduction and Application Context}
\label{introduction}

Iris segmentation is perhaps the most important step in the iris recognition pipeline. Given the complex, rich iris texture used to extract identifying features for subject matching, segmentations must correspond closely to the actual iris. Lack of precise segmentation results in misalignment of tiny iris features and, in consequence, a high probability of false matches or non-matches. In the 25 years following Daugman's seminal work on iris recognition~\cite{Daugman_TPAMI_1993}, a wide variety of approaches to iris segmentation have achieved high matching accuracy. Despite the wealth of methodologies, iris segmentation that simultaneously operates in real-time and affords a high degree of accuracy and generalization still seems to be just beyond our grasp.


The obvious tradeoff between speed and accuracy has led us to introduce a new paradigm for iris segmentation in this paper by processing iris image sequences as volumetric data. Since iris recognition cameras usually capture iris videos internally, and apply various quality checks to pick a single iris image to be used in feature extraction, we may assume that such videos are available in current commercial deployments. As a first, exploratory, step in this direction, we propose to apply the recently introduced Flexible Learning-Free Reconstruction of Neural Volumes (FLoRIN) pipeline~\cite{shahbazi2018flexible}, which was originally designed for processing volumetric image stacks of neural microscopy data. FLoRIN allows us to use information about the iris location not only in 2D space but also along the temporal axis, for example across multiple frames of an iris video.

We show that when FLoRIN is used, a minimal and acceptable drop in the overall iris recognition accuracy is compensated for by iris segmentation processing times that are a factor of $3.56$ faster than deep-learning-based solutions. When compared to other learning-free state-of-the-art iris segmentation techniques, FLoRIN-based processing is approximately an order of magnitude faster. This performance is maintained on low-resource hardware, making FLoRIN a good candidate for embedded iris matching solutions. Moreover, the FLoRIN-based iris segmentation does not need training, which is required for deep-learning-based methods, instead requiring tuning two threshold values over the interval $[0, 1]$. Following publication, we will make both the FLoRIN-based iris segmentation software and the new database of iris videos publicly available.

Summarizing, the novel contributions of this paper are:

\begin{enumerate}
    \item The application of the FLoRIN framework for volumetric iris segmentation, along with the source code of the resulting FLoRIN-based volumetric and learning-free iris segmentation tool (presented in Sec.~\ref{florin}).
    \item A novel database of iris near-infrared videos, presenting various pupil sizes as a result of visible light stimuli (presented in Sec.~\ref{pupil-dynamics}).
    \item Evaluations of FLoRIN-based, deep learning-based, and Daugman's segmentation approaches on a novel benchmark in terms of the matching accuracy and speed (described in Sec.~\ref{results}).
    \item An evaluation of FLoRIN-based and Daugman's segmentation implementations for embedded subject matching on a Raspberry Pi (presented in Sec.~\ref{results})
\end{enumerate}


\section{Related Work}
\label{related-work}

As the first step of iris recognition, iris segmentation has been approached in a variety of ways. Early and, for many years, dominant methods relied upon the circular structure of both the pupil and the iris, as proposed by Daugman \cite{Daugman_TPAMI_1993}. In addition to circular approximations of iris boundaries, those methods detected  eyelids, eyelashes, specular reflections, and excluded these occlusions from feature matching.

Subsequent approaches departed from circular approximations, and in many cases were inspired by Daugman's suggestion to use a Fourier series to provide a more complex model of the iris boundary \cite{Daugman_TSMCB_2007}. Arvacheh and Tizhoosh \cite{arvacheh2006segmentation} proposed using active contour models to detect the pupillary and limbic boundaries of the iris, taking into consideration the fact that neither the pupil nor the iris are perfectly circular, but are near-circular. They also introduced an algorithm that iteratively detects and excludes eyelid areas that cover the iris. In another approach, Shah and Ross \cite{shah2009iris} proposed a method for extracting the iris using geodesic active contours to locate the boundaries of the iris in a given image. They focused on finding the boundaries of the iris, taking into consideration the fact that the iris is not necessarily circular due to the presence of these occlusions.

Other methods approached iris segmentation from a localization perspective. Luengo \textit{et al.}~\cite{luengo2009robust} proposed using mathematical morphology to extract the outer boundaries of both the pupil and the iris. They dealt with occlusions by simply removing portions of the segmentation which likely contained parts of the eyelid, eyelashes, reflections, etc. In another approach, He \textit{et al.}~\cite{he2009toward} extracted the iris center from the image and from that detected the iris boundaries. The novelty of their algorithm was the use of a histogram filter to detect irregularities in the shape of the eyelids.

More recently, iris segmentation has moved away from these classical approaches toward machine learning oriented methods. Li \textit{et al.}~\cite{li2010robust} proposed a method which uses k-means clustering to detect the outer boundary of the iris. In a similar fashion, Sahmoud and Abuhaiba \cite{sahmoud2013efficient} proposed using k-means clustering to determine the region of the iris.

Recent advances in deep learning-based segmentation have resulted in many applications of convolutional neural network architectures to iris segmentation. Jalilian and Uhl \cite{Jalilian_DLinBiometrics_2017} proposed to use several types of convolutional encoder-decoder networks and reported better performance for deep learning-based approaches when compared to conventional algorithms such as Daugman's approach \cite{osiris}, WAHET \cite{Uhl_ICB_2012}, CAHT \cite{Rathgeb_IB_2013} and IFPP \cite{Uhl_IAR_2012}. Arsalan \etal proposed to use a modified VGG-Face network to segment visible-light \cite{Arsalan_Symmetry_2017} and near-infrared \cite{Arsalan_Sensors_2018} iris images. Other deep learning-based iris segmentation techniques include using a re-trained U-Net \etal \cite{Lozej_IWOBI_2018}, semi-parallel Fully Convolutional Deep Neural Networks \cite{bazrafkan2018end}, Generative Adversarial Networks \cite{Bezerra_SIBGRAPI_2018} and Mask R-CNN \cite{Ahmad_ArXiv_2018}. SegNet~\cite{SegNet2016} was successfully re-trained by Trokielewicz \etal \cite{TrokielewiczIWBF2018} to segment very challenging post-mortem iris images.

The ideal segmentation method would be fast, generalizable across subjects and sensors, and explainable to enable error detection and correction. Learning-based solutions are promising, however long training times and bias toward training datasets limit the performance of standard machine learning and deep learning models in the wild. Moreover, deep learning models are black boxes wherein errors are difficult to explain. By contrast, the FLoRIN pipeline used in this work is a deterministic set of operations that requires only setting two threshold values. Moreover, as will be shown, it operates on sequential image data more quickly than standard methods, enabling real-time iris segmentation. Last but not least, the number of open-sourced iris segmentation tools is currently very limited, despite a large number of papers proposing various solutions. FLoRIN-based iris segmentation software is offered with this paper.

\section{The FLoRIN Framework}
\label{florin}

\begin{figure*}[ht]
    \centering
    \includegraphics[width=\linewidth]{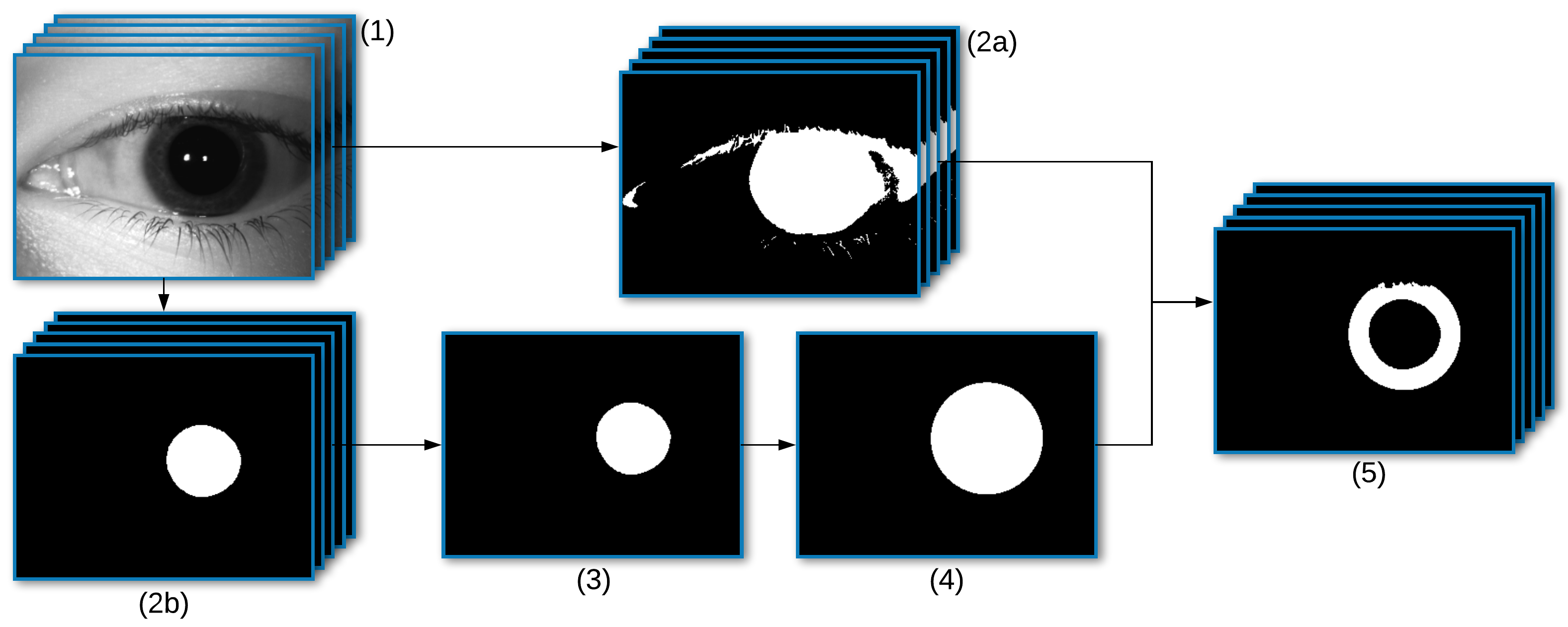}
    \caption{Overview of the FLoRIN pipeline for iris segmentation. (1) A batch of images are loaded into the pipeline. (2a) The batch is thresholded with N-Dimensional Neighborhood Thresholding (NDNT) at a low threshold value to capture the iris, pupil, and eyelids. (2b) The batch is thresholded again at a high threshold value to isolate the pupil. (3) The pupil is isolated from any other connected components and collapsed into a single image. (4) A circular mask with a diameter of half the smaller dimension of the collapsed pupil image is centered on the pupil centroid. (5) The final masks are created from the XOR of (2a) and (2b), then all pixels outside of the circular mask from (4) are zeroed. The result is a batch of masks the same size as the input batch.}
    \label{fig:pipeline}
\end{figure*}

The FLoRIN framework is a multi-stage pipeline with flexible image processing steps in each stage. FLoRIN was originally developed to meet the challenges of segmenting volumes of neural microscopy, enabling automatic discovery of non-standard structures like cells, vasculature, and myelinated axons. By incorporating volumetric data into the segmentation process through the N-Dimensional Neighborhood Thresholding (NDNT, Section~\ref{florin-ndnt}) algorithm~\cite{shahbazi2018flexible}, FLoRIN was able to boost the signal of features of interest without requiring any machine learning. In this way, FLoRIN achieved state of the art results across a number of imaging modalities in neural microscopy.

Conventionally, the FLoRIN pipeline consists of three major stages:

\begin{enumerate}
    \item \textbf{Segmentation}. The images are loaded into FLoRIN and optionally processed to improve contrast, for example by histogram equalization, Weiner filters, etc. The images are then thresholded in 2D or 3D using NDNT ~\cite{shahbazi2018flexible} parameterized by a threshold value in range [0, 1] and a pixel neighborhood size. The neighborhood size can specify a neighborhood in two or three dimensions, the latter enabling the incorporation of volumetric data into the thresholding. The binarized NDNT output is then passed to the next stage.
    \item \textbf{Identification}. The Segmentation stage output is processed to remove holes and artifacts, and morphological operations may be applied in 2D or 3D to refine the segmentation. Connected components are discovered and grouped by geometric and grayscale properties. This form of weak classification enables labeling different classes of structure; in neural microscopy this groups connected components corresponding to neurons, axons, etc. In iris segmentation, we use this stage to identify the pupil for separation from the iris pixels.
    \item \textbf{Reconstruction}. The grouped connected components are labeled and output into image files for analysis. Properties of the connected components are saved to file, from which statistics about the discovered structures may be computed. In the case of iris segmentation, the Reconstruction stage outputs binary masks corresponding to the discovered iris pixels.
\end{enumerate}

FLoRIN can operate on 2D images or 3D volumes, including time series or video data, at all stages. In particular, 3D segmentation and identification allows FLoRIN to account for volumetric data. For iris segmentation, we apply the 3D FLoRIN pipeline to videos of irises, wherein the volumetric dimension is time. This allows for the segmentation of the iris and pupil as features spanning multiple images, increasing the probability of segmenting the features of interest. In this work, we focused on adapting the Segmentation and Identification stages of FLoRIN, tailoring the pipeline to process iris images as shown in Figure~\ref{fig:pipeline}. Algorithm~\ref{alg:pipeline} provides a full description of the pipeline.

\begin{algorithm}[t]
\SetAlgoLined
\KwData{$video$: the video to process.}
\KwData{$depth$: the number of frames to process at once.}
\KwData{$t_{iris}$: iris segmentation threshold value.}
\KwData{$t_{pupil}$: pupil segmentation threshold value.}
\KwData{$w_{iris}$: iris segmentation neighborhood size.}
\KwData{$w_{pupil}$: pupil segmentation neighborhood size.}
\KwResult{A 3D array containing the iris masks.}

Create an empty array $seg$. \\

\ForEach{$block$ of size $depth$ in $video$}{
    Segment $block$ into $iris$ and $pupil$ with NDNT parameterized by $(t_{iris}, w_{iris})$ and $(t_{pupil}, w_{pupil})$. \\
    Fill holes in $iris$ and $pupil$. \\
    Remove small connected components from $iris$ and $pupil$. \\
    Create $final$ as the XOR of $iris$ and $pupil$. \\
    Collapse $pupil$ and find connected components. \\
    Create a circle centered on the detected pupil. \\
    Zero out all elements outside of the circle in each frame of $final$. \\
    Concatenate $seg$ and $final$.
}

\Return{$seg$}

\caption{FLoRIN Iris Segmentation Pipeline}
\label{alg:pipeline}
\end{algorithm}

\subsection{N-Dimensional Neighborhood Thresholding}
\label{florin-ndnt}

Central to FLoRIN is the NDNT algorithm, which is applied during the Segmentation stage to binarize images and volumes. NDNT is based on the thresholding method introduced by Bradley and Roth~\cite{bradley2007adaptive}, which uses the integral image to compute neighborhood statistics around each pixel and threshold based on the neighborhood average pixel intensity. This method was extended by Shahbazi \textit{et al.}~\cite{shahbazi2018flexible} to operate in \textit{n} dimensions, allowing for the method to account for volumetric and channel-level information.

NDNT was shown to outperform standard thresholding methods when operating on neural microscopy data, generalizing to multiple imaging modalities and features. Moreover, NDNT operates solely on the pixel neighborhood histograms, requiring tuning a single thresholding parameter in range $[0, 1]$. Compared to learning-based methods, NDNT bypasses long training times and is explainable based on the input images and neighborhood size. In the case of iris segmentation, it can account for sequential images of the same iris, strengthening the signal for segmentation.

\section{Baseline Iris Segmentation}
\label{baselines}
In this section we describe the two baseline iris segmentation methods --- SegNet~\cite{SegNet2016} and OSIRIS~\cite{osiris} --- that we compare against, as well as the subject matching pipeline used to evaluate all segmentation methods.

\subsection{SegNet DCNN Model}
\label{baselines-segnet}
To evaluate the accuracy of the proposed iris-adapted FLoRIN pipeline in comparison with a state-of-the-art segmentation method, we deploy a SegNet model \cite{SegNet2016} trained on several standard iris biometrics datasets their corresponding ground truth masks. SegNet is a fully convolutional encoder-decoder architecture, with the encoder stage being a modified VGG-16 model with removed fully connected layers, and the decoder stage comprising convolutional and upsampling layers corresponding to the max pooling layers of the encoding stage. SegNet has achieved state of the art performance on semantic segmentation tasks, including a recent successful use in segmentation of challenging post-mortem iris images \cite{TrokielewiczIWBF2018}.

\subsection{Training Procedure and Data}
To tune the original SegNet model to segment iris images, we performed a re-training procedure with images drawn from several publicly available databases, including the Biosec baseline corpus \cite{biosec} (1200 images), the BATH database\footnote{\scriptsize\url{http://www.bath.ac.uk/elec-eng/research/sipg/irisweb/}} \cite{bath} (148 images), the ND0405 database\footnote{\label{cvrl}\scriptsize\url{https://cvrl.nd.edu/projects/data/}} (1283 images), the UBIRIS database \cite{ubiris} (1923 images), and the CASIA-V4-Iris-Interval database\footnote{\scriptsize\url{http://www.cbsr.ia.ac.cn/english/IrisDatabase.asp}} (2639 images). The ground truth masks came from the IRISSEG-CC dataset by Halmstad University \cite{halmstadGT} (for the Biosec set), from the IRISSEG-EP dataset by the University of Salzburg \cite{salzburgGT} (CASIA-V4-Iris-Interval, Notre Dame 0405, and UBIRIS). For the BATH subset, we created the binary masks ourselves. The model was trained in MATLAB for 120 epochs with a batch size of 4, using an SGDM optimizer with a momentum of 0.9, learning rate of 0.001, and L2 regularization of 0.0005. The training examples were shuffled after each epoch. In total, training took approximately one day on an NVIDIA GTX 1070 GPU.

\begin{figure*}[!htbp]
    \centering
        \includegraphics[width=\textwidth]{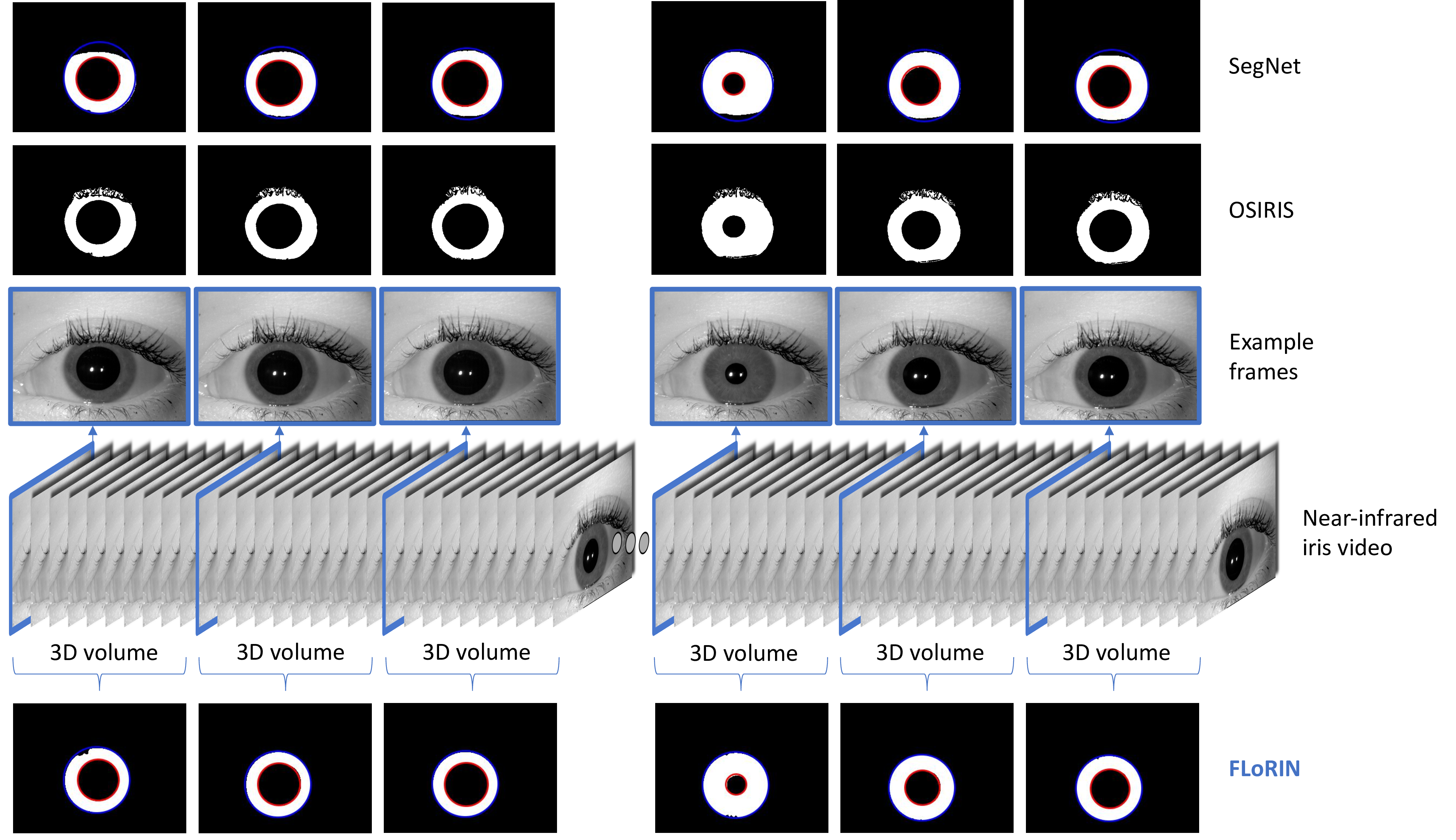}
    \caption{An example iris video recorded for this paper with selected 3D volumes processed by FLoRIN, and selected single frames processed by OSIRIS and SegNet. Red and blue circles shown in SegNet and OSIRIS segmentation masks illustrate the result of Hough transform-based curve fitting, used in post-segmentation steps in iris image normalization and matching.}
    \label{fig:dataSample}
\end{figure*}

\subsection{OSIRIS Segmentation and Matching}
\label{baselines-osiris}

In addition to the SegNet model, we also evaluate the original, unmodified OSIRIS segmentation. OSIRIS is an open-source academic matcher, which implements the principles of original Daugman concept, including iris segmentation using the circular Hough transform and subsequent refinement with active contours, iris image normalization, and Gabor-based filtering, which yields binary codes that allow for efficient matching. In this paper, we use OSIRIS in three different modes:
\begin{enumerate}
    \item {\bf baseline conventional:} stock OSIRIS for both segmentation, matching, and encoding
    \item {\bf baseline DCNN:} SegNet masks and OSIRIS encoding and matching
    \item {\bf proposed:} FLoRIN masks and OSIRIS encoding and matching
\end{enumerate}
Each mode allows us to obtain ROC (Receiver Operating Characteristic) curves, which we use to denote the recognition accuracy that each set of iris masks enables.

\section{Benchmark Iris Video Dataset}
\label{pupil-dynamics}

To enable processing volumetric data with FLoRIN, we used a newly collected database of iris videos, acquired in near-infrared, which is released as a part of this work. To include a wide spectrum of pupil dilation, we followed the acquisition protocol defined by Czajka \cite{Czajka_TIFS_2015}, and collected 30-second videos, at a rate of 25 frames per second, of an iris stimulated by visible light. That is, during the first 15 seconds, 375 iris images are taken in darkness (yet small pupil oscillations, called {\it hippus} are still present). During the next 5 seconds, 125 iris images are taken right after the visible light is switched on. This forces the pupil's rapid constriction and acquisition of sample under varying pupil size. The last 250 images are taken right after the visible light is switched off, making the pupil to dilate and providing additional samples with varying size of the iris texture. Hence, each video comprises 750 iris images. All samples are compliant with ISO/IEC 19794-6 standard. Fig. \ref{fig:dataSample} presents selected frames taken from an example video, along with OSIRIS, SegNet and FLoRIN segmentations.

42 subjects participated in the data acquisition. We collected from one to four videos for each subject at the 25 FPS rate. Except for 17 videos, all sequences comprise 750 frames, and the total number of iris images considered in this work is 117,717. Researchers interested in obtaining a copy of this dataset are requested to follow the instructions provided at \url{http://...}\footnote{link removed temporarily to make this submission anonymous}.


\section{Experiments}
\label{results}

A key question we sought to answer in our experiments is: how close to top performing iris segmentation performance can one get with a learning-free method that is optimized for speed? To answer this question, we processed the collected iris video data with the proposed FLoRIN pipeline (Section~\ref{florin}. We then compared FLoRIN matching performance with segmentations generated by SegNet and OSIRIS (Section~\ref{baselines-segnet}), both of which are limited to 2D processing. All code used in these evaluations will be released upon publication.

For the assessment of the matching accuracy, we normalize the segmentation outputs produced by the SegNet and FLoRIN algorithms to fit them into the OSIRIS recognition pipeline (Section~\ref{baselines-osiris}. Circular Hough Transform (CHT) is employed to approximate the iris and pupil boundaries in the obtained binary masks. OSIRIS to normalizes the iris images and corresponding masks onto a dimensionless polar-coordinate rectangle, which is then used in the Gabor filtering and encoding stage.

Genuine matching performance was computed by first selecting every tenth frame of each video, then comparing the selected frames. Impostor matching performance was computed by first selecting every twentieth frame of the first video for each eye, then comparing the selected frames representing different eyes. Approximately 100k genuine comparisons and approximately 200k impostor comparisons were conducted. Comparison pairs were identical for all tested methods.

Our analysis shows that FLoRIN dramatically increases the speed of segmentation while incurring only minor penalties to subject matching performance. For each video, FLoRIN processed batches of 5 frames at a time. All throughput analyses were computed on a system with 4 Intel Xeon E5-2650 v4 processors, 24GB of RAM, and an NVIDIA GTX 1080ti GPU, using versions of the videos downsampled to $320 \times 240$ pixels per frame.

\begin{table}[!htbp]
\centering
\begin{tabular}{| c | c | c | c | }
 \hline
 {\textbf{Method}}                    &  {\textbf{AUC}} & {\textbf{EER}} & {\textbf{Mean FPS}} \\
 \hline
 {FLoRIN~\cite{shahbazi2018flexible}} & 0.962 & 0.074 & \textbf{37.38 $\pm$ 2.58} \\
 \hline
 {SegNet~\cite{SegNet2016}}           & 0.992 & \textbf{0.016} & $10.50 \pm 1.14$ \\
 \hline
 {OSIRIS~\cite{osiris}}               & \textbf{0.996} & 0.017 & $\:\,3.24 \pm 0.28$ \\
 \hline
\end{tabular}
\caption{Matching Performance and Segmentation Throughput. The mean frames per second (FPS) is computed as the average per-video FPS over all 159 videos.}
\label{tab:perf}

\end{table}

FLoRIN offers a substantial increase in throughput over SegNet and OSIRIS. As shown in Table~\ref{tab:perf}, FLoRIN is a factor of 3.56 faster than SegNet and an order of magnitude faster than OSIRIS processing the same videos. This throughput increase is a direct result of processing videos in 3D: a batch of frames is segmented in parallel using surrounding frames to boost the signal, then all frames in the batch are post-processed simultaneously to improve the segmentation. This scheme introduces data parallelism across the time dimension of the videos which is unavailable to SegNet, OSIRIS, and other 2D segmentation methods.

\begin{figure}[!tbp]
    \centering
    \includegraphics[width=\columnwidth]{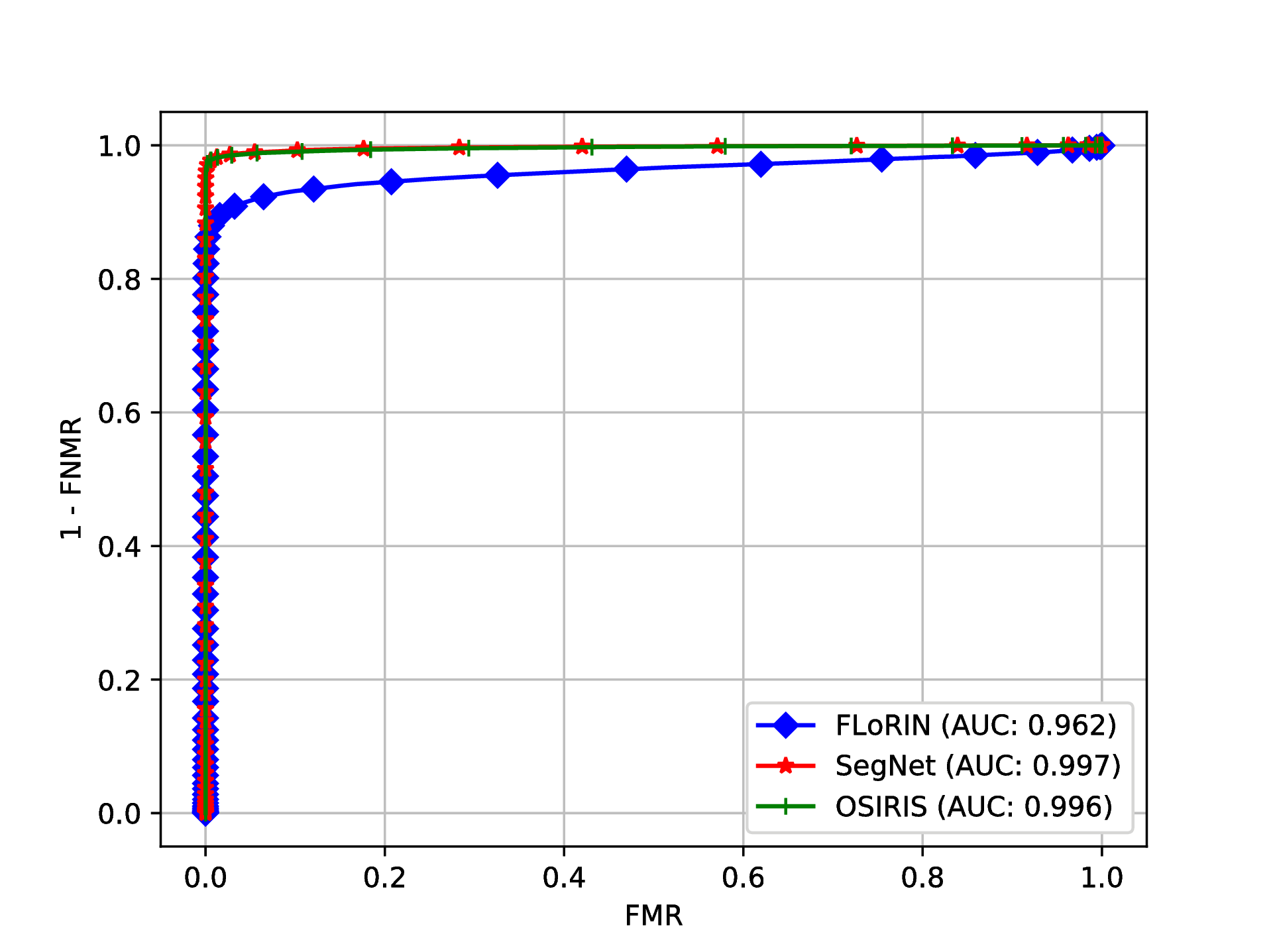}
    \caption{Receiver operating characteristic of the False Match Rate (FMR) versus the inverse False Non-Match Rate (FNMR) for FLoRIN, SegNet, and OSIRIS on the pupil dynamics dataset.}
    \label{fig:roc}
\end{figure}

The increased speed of FLoRIN comes with minor subject matching performance degradation. The receiver operating characteristic (ROC) curve of the inverse False Negative Match Rate vs the False Match Rate of subject matching for each method is shown in Figure~\ref{fig:roc}. FLoRIN has an area under the curve (AUC) of 0.96 and an equal error rate (EER) of 0.07 (Table~\ref{tab:perf}), while SegNet and OSIRIS both boast an AUC of 0.99 and an EER of 0.02. Qualitatively, we find that the FLoRIN segmentations tend to under-segment the pupil, leading to a greater proportion of non-iris pixels included in the normalized masks. This issue will be addressed with additional quality control operations as FLoRIN continues to develop.

\begin{figure*}
    \centering
    \begin{subfigure}[t]{0.19\textwidth}
        \includegraphics[width=\textwidth]{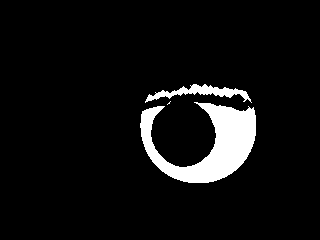}
        \caption{48\_L\_1 Frame 286}
    \end{subfigure}
    \begin{subfigure}[t]{0.19\textwidth}
        \includegraphics[width=\textwidth]{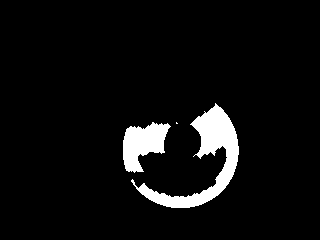}
        \caption{53\_L\_1 Frame 132}
    \end{subfigure}
    \begin{subfigure}[t]{0.19\textwidth}
        \includegraphics[width=\textwidth]{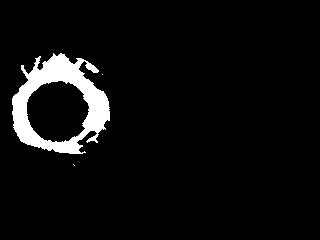}
        \caption{62\_L\_1 Frame 27}
    \end{subfigure}
    \begin{subfigure}[t]{0.19\textwidth}
        \includegraphics[width=\textwidth]{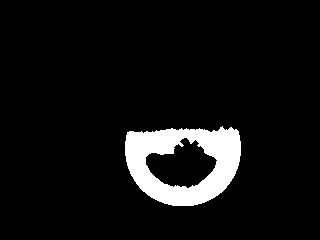}
        \caption{73\_R\_1 Frame 223}
    \end{subfigure}
    \begin{subfigure}[t]{0.19\textwidth}
        \includegraphics[width=\textwidth]{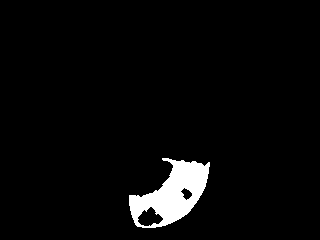}
        \caption{85\_L\_1 Frame 14}
    \end{subfigure}
    \caption{Examples of lower-quality FLoRIN segmentations. These segmentations are the result of mis-identification of the pupil connected component, remaining pixels connecting the pupil and eyelid, and blinking on the part of the subject. The quality of these segmentations can be improved with per-batch threshold parameters and quality control mechanisms to remove blink frames.}
    \label{fig:badsegs}
\end{figure*}

Throughout this experiment, FLoRIN thresholding parameters were manually selected on a per-video basis, however in many cases the threshold values transferred between videos of the same subject. Parameter values were selected by a sweep over the real-valued domain $[0, 1]$ applied to a subset of each video. The results were then manually examined to determine the optimal threshold values. By reusing intermediate data computed by the NDNT algorithm that is invariant to threshold value, we were able to quickly segment these subsets across a large number of threshold values. We evaluated this scheme across subsets of 5 frames of each video with threshold values spaced at $10^{-2}$ and found that the parameter sweep could be completed in approximately $2$s for a given window size.

\subsection{Recommendations for Threshold Parameters}
\label{results-tvals}

While manually tuning the iris and pupil threshold values for each of the videos, we discovered a number of trends in the selection of threshold values and neighborhood windows sizes. We break these recommendations down by feature to enable others to apply FLoRIN to new data.

\textbf{Iris Parameters.} As the iris spans a large percentage of each frame in a video, we found that a larger neighborhood was better able to capture the iris pixels. In every video, we used a window of size $2\times 256\times 256$ around each pixel, which accounts for the previous and subsequent frames. Given the size of the window, we recognize that this is similar to a global thresholding method with the inclusion of volumetric data. Threshold values for the iris tended to the domain $[0.2, 0.5]$, with a sensitivity of $0.01$.

\textbf{Pupil Parameters.} Pupil segmentation was more sensitive to neighborhood size, relative to the shape of the image histogram. In the case of a histogram tending toward the right, a large $2\times 256\times 256$ neighborhood with threshold values in domain $[0.7, 0.95]$ sufficed to capture the pupil. In extreme cases with a tall peak of low-intensity values, for example when the subject wore eyeliner, we found that $2\times 64\times 64$ or $2\times 32\times 32$ neighborhoods were better suited to isolate the pupil. The domain of threshold values differs for smaller neighborhood sizes, typically between $[0.1\mbox{--}0.3]$, with a sensitivity of $0.01$.

In both cases, the threshold values could be narrowed to a subset of $[0, 1]$. These findings indicate that the optimal threshold value is a function of the image histogram and the window size. This function could be used to develop an automatic parameterization scheme, either through machine learning or another optimization scheme.

\subsection{Performance on Low-Resource Hardware}
\label{results-pi}

Given the large improvement in running time performance of FLoRIN over SegNet and OSIRIS, Table~\ref{tab:perf}, we hypothesized that FLoRIN would be suitable for low-resource hardware and embedded systems. We conducted a second timing experiment on a Raspberry Pi 3 with 4 cores and 1GB of RAM. We processed the first 20 frames of each video on the Pi with FLoRIN and OSIRIS using the same setup used to process the full videos.

With the limited resources available to the Pi, FLoRIN processed the video subset with a mean throughput of $4.50 \pm 0.36$ frames per second per video. This was an order of magnitude greater than OSIRIS, which had a throughput of $0.38 \pm 0.03$ frames per second per video. The throughput afforded by FLoRIN on the Pi indicates that FLoRIN is fast enough to enable segmentation on embedded systems, e.g. commercial iris recognition hardware.

\section{Discussion}
\label{conclusion}

Deep learning has, in many ways, revolutionized the field of biometrics. However, it is not the only way to approach problems like iris segmentation. Legitimate criticisms of deep learning exist in the form of long training times, the need for large amounts of hand-labeled training data, and the complexity of optimizing various hyperparameters. And frustratingly, even when all of these problems have been addressed for a single dataset or operational scenario, the move to a different setting forces one to start all over again. Generalization remains an open problem within the field of machine learning at large.

Our turn back to learning-free methods is a direct response to this dilemma. Elaborate training regimes leaning on massive data collection and annotation efforts can be dispensed with in favor of immediate inference operation. A move to a new dataset or acquisition setting may be as simple as making a few adjustments to a minimal set of free parameters before deployment. Several decades worth of work on image processing and computer vision should not be ignored --- the literature is filled with older techniques that can be updated for today's problems to achieve remarkable performance gains and avoid the generalization dilemma. The FLoRIN approach is just one example of this.

Given the observed speed of iris segmentation with FLoRIN, we propose that the pipeline used in this work can be deployed as a semi-automated annotator for generating ground truth. Using FLoRIN, a new dataset may be processed rapidly, and the output masks may be proofread by human annotators to create pixel-level ground truth labels. This scheme will enable rapid release and dissemination of new databases for use across the biometrics community.

This study represents the initial application of FLoRIN to iris segmentation, and as such required manually tuning the parameters of the NDNT algorithm. We are in the process of evaluating a number of methods for automatic parameterization, including grid search across threshold values using Reverse Classification Accuracy~\cite{valindria2017reverse} as a guide and regression models to map images to threshold values. Such automation will reduce the need for a human in the loop. Based on the results of Section~\ref{results-pi}, we believe an automatically-parameterized FLoRIN will be ideal for embedded iris biometrics systems.


\bibliographystyle{ieee}
\bibliography{florin}

\begin{thebibliography}{10}\itemsep=-1pt

\bibitem{Ahmad_ArXiv_2018}
S.~{Ahmad} and B.~{Fuller}.
\newblock {Unconstrained Iris Segmentation using Convolutional Neural
  Networks}.
\newblock {\em arXiv e-prints}, page arXiv:1812.08245, Dec. 2018.

\bibitem{halmstadGT}
F.~Alonso-Fernandez and J.~Bigun.
\newblock Near-infrared and visible-light periocular recognition with gabor
  features using frequency-adaptive automatic eye detection.
\newblock {\em IET Biometrics}, 4(2):74--89, 2015.

\bibitem{Arsalan_Symmetry_2017}
M.~Arsalan, H.~G. Hong, R.~A. Naqvi, M.~B. Lee, M.~C. Kim, D.~S. Kim, C.~S.
  Kim, and K.~R. Park.
\newblock Deep learning-based iris segmentation for iris recognition in visible
  light environment.
\newblock {\em Symmetry}, 9(11), 2017.

\bibitem{Arsalan_Sensors_2018}
M.~Arsalan, R.~A. Naqvi, D.~S. Kim, P.~H. Nguyen, M.~Owais, and K.~R. Park.
\newblock Irisdensenet: Robust iris segmentation using densely connected fully
  convolutional networks in the images by visible light and near-infrared light
  camera sensors.
\newblock {\em Sensors}, 18(5), 2018.

\bibitem{arvacheh2006segmentation}
E.~Arvacheh and H.~Tizhoosh.
\newblock Iris segmentation: Detecting pupil, limbus and eyelids.
\newblock In {\em IEEE ICIP}, pages 2453--2456, 2006.

\bibitem{SegNet2016}
V.~Badrinarayanan, A.~Kendall, and R.~Cipolla.
\newblock {SegNet: A Deep Convolutional Encoder-Decoder Architecture for Image
  Segmentation}.
\newblock {\em IEEE T-PAMI}, 39(12), 2017.

\bibitem{bazrafkan2018end}
S.~Bazrafkan, S.~Thavalengal, and P.~Corcoran.
\newblock An end to end deep neural network for iris segmentation in
  unconstrained scenarios.
\newblock {\em Neural Networks}, 106:79--95, 2018.

\bibitem{Bezerra_SIBGRAPI_2018}
C.~S. Bezerra, R.~Laroca, D.~R. Lucio, E.~Severo, L.~F. Oliveira, A.~S. Britto,
  and D.~Menotti.
\newblock Robust iris segmentation based on fully convolutional networks and
  generative adversarial networks.
\newblock In {\em SIBGRAPI}, 2018.

\bibitem{bradley2007adaptive}
D.~Bradley and G.~Roth.
\newblock Adaptive thresholding using the integral image.
\newblock {\em Journal of Graphics Tools}, 12(2):13--21, 2007.

\bibitem{Czajka_TIFS_2015}
A.~Czajka.
\newblock Pupil dynamics for iris liveness detection.
\newblock {\em IEEE T-IFS}, 10(4):726--735, April 2015.

\bibitem{Daugman_TPAMI_1993}
J.~Daugman.
\newblock High confidence visual recognition of persons by a test of
  statistical independence.
\newblock {\em IEEE T-PAMI}, 15(11):1148--1161, Nov 1993.

\bibitem{Daugman_TSMCB_2007}
J.~Daugman.
\newblock New methods in iris recognition.
\newblock {\em IEEE T-SMC, Part B}, 37(5):1167--1175, Oct 2007.

\bibitem{biosec}
J.~Fierrez, J.~Ortega-Garcia, D.~T. Toledano, and J.~Gonzalez-Rodriguez.
\newblock Biosec baseline corpus: A multimodal biometric database.
\newblock {\em Pattern Recognition}, 40(4):1389 -- 1392, 2007.

\bibitem{he2009toward}
Z.~He, T.~Tan, Z.~Sun, and X.~Qiu.
\newblock Toward accurate and fast iris segmentation for iris biometrics.
\newblock {\em IEEE T-PAMI}, 31(9), 2009.

\bibitem{salzburgGT}
H.~Hofbauer, F.~Alonso-Fernandez, P.~Wild, J.~Bigun, and A.~Uhl.
\newblock A ground truth for iris segmentation.
\newblock In {\em ICPR}, August 2014.

\bibitem{Jalilian_DLinBiometrics_2017}
E.~Jalilian and A.~Uhl.
\newblock {\em Iris Segmentation Using Fully Convolutional Encoder--Decoder
  Networks}, pages 133--155.
\newblock Springer International Publishing, Cham, 2017.

\bibitem{li2010robust}
P.~Li, X.~Liu, L.~Xiao, and Q.~Song.
\newblock Robust and accurate iris segmentation in very noisy iris images.
\newblock {\em IVC}, 28:246--253, 2010.

\bibitem{Lozej_IWOBI_2018}
J.~Lozej, B.~Meden, V.~Struc, and P.~Peer.
\newblock End-to-end iris segmentation using {U-Net}.
\newblock In {\em IWOBI}, July 2018.

\bibitem{luengo2009robust}
M.~A. Luengo-Oroz, E.~Faure, and J.~Angulo.
\newblock Robust iris segmentation on uncalibrated noisy images using
  mathematical morphology.
\newblock {\em IVC}, 28:278--284, 2009.

\bibitem{bath}
D.~Monro, S.~Rakshit, and D.~Zhang.
\newblock {University of Bath, UK Iris Image Database}, 2009.

\bibitem{osiris}
N.~Othman, B.~Dorizzi, and S.~Garcia-Salicetti.
\newblock {OSIRIS: An open source iris recognition software}.
\newblock {\em Pattern Recognition Letters}, 82(2):124--131, 2016.

\bibitem{ubiris}
H.~Proenca, S.~Filipe, R.~Santos, J.~Oliveira, and L.~Alexandre.
\newblock The {UBIRIS.v2}: A database of visible wavelength images captured
  on-the-move and at-a-distance.
\newblock {\em IEEE T-PAMI}, 32(8):1529--1535, August 2010.

\bibitem{Rathgeb_IB_2013}
C.~Rathgeb.
\newblock Iris segmentation methodologies.
\newblock In C.~Rathgeb, A.~Uhl, and P.~Wild, editors, {\em Iris biometrics:
  From segmentation to template security. In Advances in Information Security}.
  Springer, 2013.

\bibitem{sahmoud2013efficient}
S.~A. Sahmoud and I.~S. Abuhaiba.
\newblock Efficient iris segmentation method in unconstrained environments.
\newblock {\em Pattern Recognition}, 46:3174--3185, 2013.

\bibitem{shah2009iris}
S.~Shah and A.~Ross.
\newblock iris segmentation using geodesic active contours.
\newblock {\em IEEE T-IFS}, 4(4):824--836, 2009.

\bibitem{shahbazi2018flexible}
A.~Shahbazi, J.~Kinnison, R.~Vescovi, M.~Du, R.~Hill, M.~Joesch, M.~Takeno,
  H.~Zeng, N.~M. da~Costa, J.~Grutzendler, et~al.
\newblock Flexible learning-free segmentation and reconstruction of neural
  volumes.
\newblock {\em Scientific Reports}, 8(1):14247, 2018.

\bibitem{TrokielewiczIWBF2018}
M.~Trokielewicz and A.~Czajka.
\newblock Data-driven segmentation of post-mortem iris images.
\newblock In {\em IWBF}, June 2018.

\bibitem{Uhl_IAR_2012}
A.~Uhl and P.~Wild.
\newblock Multi-stage visible wavelength and near infrared iris segmentation
  framework.
\newblock In A.~Campilho and M.~Kamel, editors, {\em Image Analysis and
  Recognition}, pages 1--10, Berlin, Heidelberg, 2012. Springer Berlin
  Heidelberg.

\bibitem{Uhl_ICB_2012}
A.~Uhl and P.~Wild.
\newblock Weighted adaptive hough and ellipsopolar transforms for real-time
  iris segmentation.
\newblock In {\em ICB}, March 2012.

\bibitem{valindria2017reverse}
V.~V. Valindria, I.~Lavdas, W.~Bai, K.~Kamnitsas, E.~O. Aboagye, A.~G. Rockall,
  D.~Rueckert, and B.~Glocker.
\newblock Reverse classification accuracy: predicting segmentation performance
  in the absence of ground truth.
\newblock {\em IEEE T-MI}, 36(8):1597--1606, 2017.

\end{thebibliography}

\end{document}